\documentclass[conference]{IEEEtran}
\IEEEoverridecommandlockouts
\usepackage{cite}
\usepackage{amsmath,amssymb,amsfonts}
\usepackage{algorithmic}
\usepackage{graphicx}
\usepackage{textcomp}
\usepackage{xcolor}
\usepackage{multirow}
\usepackage{tabularx}

\usepackage{xspace}
\usepackage{marginnote}
\newtheorem{lemma}{Lemma}
\newtheorem{example}{Example}
\newtheorem{definition}{Definition}

\newcommand{\simulator}{\mathcal{E}}

\newcommand{\ego}{\mathrm{ego}}
\newcommand{\NPC}{\mathrm{NPC}}
\newcommand{\commentout}[1]{}
\ifCLASSOPTIONcompsoc
    \usepackage[caption=false,font=normalsize,labelfont=sf,textfont=sf]{subfig}
\else
    \usepackage[caption=false,font=footnotesize]{subfig}
\fi
\usepackage{booktabs}
\usepackage{algorithm}
\usepackage{algorithmic}
\usepackage{array}

\def\BibTeX{{\rm B\kern-.05em{\sc i\kern-.025em b}\kern-.08em
    T\kern-.1667em\lower.7ex\hbox{E}\kern-.125emX}}
\begin{document}

\title{Safety Analysis of Autonomous Driving Systems Based on Model Learning
}

% \author{Anonymity}

\commentout{
\author{\IEEEauthorblockN{1\textsuperscript{st} Renjue Li}
\IEEEauthorblockA{\textit{SKLCS, Institute of Software, CAS} \\
\textit{University of Chinese Academy of Sciences}\\
Beijing, China \\
lirj19@ios.ac.cn}
\and
\IEEEauthorblockN{2\textsuperscript{nd} Tianhang Qin}
\IEEEauthorblockA{\textit{SKLCS, Institute of Software, CAS} \\
\textit{University of Chinese Academy of Sciences}\\
Beijing, China \\
qinht@ios.ac.cn}
\and
\IEEEauthorblockN{3\textsuperscript{rd} Pengfei Yang}
\IEEEauthorblockA{\textit{SKLCS, Institute of Software, CAS} \\
\textit{University of Chinese Academy of Sciences}\\
Beijing, China \\
yangpf@ios.ac.cn}
\and
\IEEEauthorblockN{4\textsuperscript{th} Cheng-Chao Huang}
\IEEEauthorblockA{\textit{Nanjing Institute of Software Technology, ISCAS} \\
\textit{Pazhou Lab}\\
Beijing, China \\
chengchao@nj.iscas.ac.cn}
\and
\IEEEauthorblockN{5\textsuperscript{th} Youcheng Sun}
\IEEEauthorblockA{\textit{Queen's University Belfast} \\
\textit{name of organization (of Aff.)}\\
Manchester, United Kingdom \\
youcheng.sun@qub.ac.uk}
\and
\IEEEauthorblockN{6\textsuperscript{th} Lijun Zhang}
\IEEEauthorblockA{\textit{SKLCS, Institute of Software, CAS} \\
\textit{University of Chinese Academy of Sciences}\\
Beijing, China \\
zhanglj@ios.ac.cn}
}
}

\author{\IEEEauthorblockN{Renjue Li \IEEEauthorrefmark{2},
Tianhang Qin\IEEEauthorrefmark{2}, Pengfei Yang\IEEEauthorrefmark{2},
Cheng-Chao Huang \IEEEauthorrefmark{3},
Youcheng Sun \IEEEauthorrefmark{4}, and Lijun Zhang \IEEEauthorrefmark{2}}
\IEEEauthorblockA{\textit{SKLCS, Institute of Software, CAS}\IEEEauthorrefmark{2} \\
\textit{University of Chinese Academy of Sciences}\IEEEauthorrefmark{2}\\
\textit{Nanjing Institute of Software Technology, ISCAS}\IEEEauthorrefmark{3}\\
\textit{The University of Manchester}\IEEEauthorrefmark{4}\\
Emails: \{lirj19, qinht, yangpf, zhanglj\}@ios.ac.cn,
chengchao@nj.iscas.ac.cn,
youcheng.sun@manchester.ac.uk}}

\maketitle

% \begin{abstract}
% % The behaviour of an autonomous driving system (ADS) is highly dependent on the traffic scenarios it is in. 
% To analyse the safety of autonomous driving systems (ADS) in traffic scenarios, we present a practical verification framework from a model learning perspective. %Unlike traditional testing methods, 
% The main idea is to build a surrogate model that quantitatively depicts the behaviour of an ADS, subject to the specified traffic scenario.
% % The surrogate model accepts scenario parameters as its input and output a quantified value of the behaviour we care about, which is iteratively trained from the simulation results under the scenarios generated within a given parameter space.
% We train the surrogate iteratively under scenarios generated within a given parameter space and bound the absolute error with probabilistic guarantee.
% By verifying the surrogate model, we can verify whether the ADS satisfies a safety criterion for the entire parameter space with the probabilistic guarantee.
% %When the verification fails, 
% If the ADS fails the safety criterion, we can further compute the quantitative indicators to measure the association between its unsafe behaviour and each parameter. Besides, we can generate the region of unsafe parameters guided by the counter-examples we obtained.
% These will help to reveal in depth the weaknesses of the ADS.
% \end{abstract}

\begin{abstract}
% 100 words
We present a practical verification method for safety analysis of the autonomous driving system (ADS).
The main idea is to build a surrogate model that quantitatively depicts the behaviour of an ADS in the specified traffic scenario. The safety properties proved in the resulting surrogate model apply to the original ADS with a probabilistic guarantee.
Furthermore, %over/under-approximations of 
we explore the safe and the unsafe parameter space of the traffic scenario for driving hazards. 
We demonstrate the utility of the proposed approach by evaluating safety properties on the state-of-the-art ADS in literature, with a variety of simulated traffic scenarios.
%The surrogate is trained with the simulations generated within a given parameter space,
%which is utilised to verify whether the ADS satisfies a safety criterion with the probabilistic guarantee.
%When the ADS fails the safety criterion, the association between its unsafe behaviour and each parameter can be measured and the region of unsafe parameters can be inner-approximated.
%For experiment, we analyse the state-of-the-art ADS---LAV with 5 scenarios.

\end{abstract}


\section{Introduction}
Autonomous driving systems (ADS) 
%of different autonomy levels (Figure \ref{fig:safelevel}) 
are expected to bring forth an efficient and safe road traffic. %, which can help reduce the cost of transporting. 
With the great success of artificial intelligence (AI), such autonomous driving systems are designed and equipped with various  components, taking various sensor inputs and performing perception and prediction tasks. Thus, the safety guarantee of such AI enabled autonomous driving systems is a key challenge.

%To evaluate and improve the safety of an autonomous driving system, testing is a critical approach. 
In order to test autonomous vehicles in real-world scenarios, tremendous resources are required to build scenes and simulate real traffics, so it is unacceptable to thoroughly test an ADS in the real world. The development of driving simulators such as CARLA~\cite{CARLA} and BeamNG~\cite{BeamNG} dramatically reduces the testing cost by introducing a virtual simulated environment. Based on the simulator, a variety of testing approaches have been developed to generating test cases and analysing different traffic scenes \cite{fremont2019scenic}. %measurements to help the analysis of testing approaches, which can generate testing cases heuristically \cite{}. 
However, even that many unsafe testing cases have been found by these approaches, such testing approaches provide almost no safety guarantee to the ADS.

Search-based testing approaches~\cite{AbdessalemNBS18, ArcainiZI21, calo2020generating, borg2021digital, gambi2019automatically, gambi2019asfault, tian2022mosat, haq2022efficient} try to find the parameters that make the ADS misbehave in the corresponding testing scenarios. These approaches often use a fitness function, e.g. time-to-collision~\cite{timetocollision}, to guide the search process. 
%The genetic algorithm was used in~\cite{} to optimize the fitness function for possible dangerous parameters. Some work also exploits the surrogate models to accelerate the search~\cite{}. 
In our work, we adopt this idea of  fitness functions to specify safety properties. At the same time, inspired by the previous work on linear model learning from a deep neural network~\cite{deeppac}, we propose an algorithm to learn a  fully connected neural network (FNN) model to approximating the fitness function. 
%The difference is that our surrogate model has a probabilistic guarantee on its absolute difference with the ground truth. We exploits the power of adversarial attacks in deep learning and replace the genetic algorithm with the PGD algorithm. 
Different from testing based approaches, the learned FNN can be proven to be probably approximately correct (PAC), which cannot be achieved by prior ADS testing methods. 

In this paper, we focus on the formal verification for the safety properties of the ADS. 
In contrast to testing, formal verification aims to give a mathematical proof to a given property of the system with the system formally modeled and the property formally specified.
An ADS can be modeled as a neural network controlled system (NNCS).
The safety verification of NNCS based on reachability analysis has been studied in previous works using activation function reduction~\cite{ivanov2019verisig}, abstract interpretation~\cite{tran2020nnv} and function approximation~\cite{ivanov2020verifying, ivanov2021verisig, huang2019reachnn, fan2020reachnn, huang2022polar}. 
These white-box methods over-approximate the behaviour of neural networks using Tyler model arithmetic or abstract interpretation, and are too inefficient for large systems like ADS. Unlike the reachability analysis, our approach gives quantified certificate of safety properties instead, in a much more efficient way and a black-box manner that is more general.

In particular, in our work we verify the safety property of a given ADS, under various traffic scenarios, with a probabilistic guarantee. For example, with $99.9\%$ confidence, the ADS is collision-free with probability at least $99\%$ in the emergency braking scenario. A traffic scenario can comprise parameters such as vehicle velocity, weather, etc.
We are interested in whether the ADS can navigate safely in the traffic scenario with a wide range of parameters.

The idea behind our approach is to learn a surrogate model that approximates the fitness function of the original ADS with a measurable difference gap. If the surrogate model is proved to be safe in a traffic scenario, we can then derive a probabilistic guarantee on the safety property for the ADS in the same scenario.
If the surrogate model fails to meet the safety property, we further explore its parameter space, by dividing whole space into cells according to specified parameters and analysing the quantified level of safety in each of these cells. Therefore, our verification framework is quantitative in the formal specification of safety properties,  the learned surrogate model with its probabilistic guarantee, and the further analysis of safe/unsafe regions.

In summary, the contributions in this paper are at least three fold.

\begin{itemize}
    \item We purpose a practical verification framework for the scenario specific safety of an autonomous driving system with probabilistic guarantee, based on surrogate model learning. %To the best of our knowledge, this is the first verification framework for complex autonomous driving systems.
    \item %From the learned surrogate model, 
    When the verification fails, we can further conduct a quantitative analysis on the parameters of interest, and partition the parameter space into
    safe and potentially unsafe regions for the surrogate model. The analysis results are essential for improving the ADS.%the dangerous parameter values if the ADS does not meet the safe criterion. Such dangerous parameter values are analysed in terms of both inner-approximation and outer-approximation.
    \item We apply the developed verification approach and parameter space exploration method to the state-of-the-art ADS %published in CVPR~2022, i.e. LAV, 
    with five traffic scenarios in CARLA. The results confirm that the learning-based verification is a promising future direction for ADS. 
\end{itemize}

\section{Background}
\label{sect:background}

% \begin{figure}[t]
%     \centering
%         \centering
%         \includegraphics[width=1\linewidth]{figures/safeLevel.eps}
%     \caption{The levels of Autonomous Driving Systems.}
%     \label{fig:safelevel}
% \end{figure}

\subsection{Autonomous driving systems}
\label{subsect:criteria}
An autonomous driving system is designed to assist or even replace the human drivers in real traffic. By the degree of automation, ADS can be categorized into six levels, i.e. from L0 to L5\commentout{~\cite{SAEstandard}}. The desired L5 ADS can perform all actions required for all possible traffic scenarios without any help or human interactions.
A modern ADS receives the data from various sensors and generates the control signals using underlying models. A typical ADS we study has an architecture consisting sensors, perception module, prediction module, planning module and control module.

The autonomy of a vehicle must not compromise the on-road safety. In Table~\ref{tab:common_safe_criterion}, there are examples of requirements for guaranteeing that an ADS navigates safely under different traffic scenarios, in which we mainly focus on the collision free property in this paper. %\textbf{LZ: which properties in the table do we analyse, all of them? (Answered)} %many criterion has been proposed. We list some common criterion for the ADS as follow:
% \begin{itemize}
%     \item{Collision Free}: The most critical requirement for ADS is not to collide with other vehicles, etc. This criterion evaluates the general safety of an ADS by judging whether a collision is occurred.
%     \item{Route Completion}: The functionality of an ADS can be measured with its ability to navigate and control the vehicle moving from one location to another. This criterion is designed to guarantee the percentage of the route completed by a vehicle with ADS.
%     \item{Traffic Lights}: Running the red light is one of the typical traffic misbehaviour. This criterion monitors whether the developed ADS can detect and follow the traffic lights correctly under diverse conditions. 
%     \item{Vehicle Speed}: There can be speed limits for different roads. This criterion requires the autonomous vehicle keep its speed in a reasonable range.
%     \item{Lane Keeping}: To maintain the stability of an autonomous driving system, the car should always drive in traffic lanes as long as there is no overtaking or lane changing actions in progress. The ADS fails this criterion if the car ran over the lane where it should follow.
% \end{itemize}

\begin{table}[h]
\renewcommand{\arraystretch}{1.2}
\caption{Common Safety Properties for ADS}
\centering
\def\mystrut{\rule{0pt}{1.2\normalbaselineskip}}
\begin{tabularx}{0.49\textwidth}{lX}
\hline
Safety Properties  & Description                                                                                                                                                                                                                                                       \\ \hline \mystrut
Collision Free   & The most critical requirement for ADS is not to collide with other vehicles, etc. This property evaluates the general safety of an ADS by judging whether a collision is occurred.                                                                               \\ \mystrut
Route Completion & The functionality of an ADS can be measured with its ability to navigate and control the vehicle moving from one location to another. This property is designed to guarantee the percentage of the route completed by a vehicle with ADS.                        \\ \mystrut
Traffic Lights   & Running the red light is one of the typical traffic misbehaviour. This property monitors whether the developed ADS can detect and follow the traffic lights correctly under diverse conditions.                                                                   \\ \mystrut
Vehicle Speed    & There can be speed limits for different roads. This property requires the autonomous vehicle keep its speed in a reasonable range.                                                                                                                               \\ \mystrut
Lane Keeping     & To maintain the stability of an autonomous driving system, the car should always drive in traffic lanes as long as there is no overtaking or lane changing actions in progress. The ADS fails this property if the car ran over the lane where it should follow. \\ \hline
\end{tabularx}
\label{tab:common_safe_criterion}
\end{table}

\subsection{CARLA \& Scenairo Runner}
We use the high-fidelity simulator CARLA to generate the traffic scenarios. CARLA is based on Unreal Engine 4\commentout{~\cite{unrealengine}}, and it supports the real-time simulation of sensors, dynamic physics and traffic environments. It also provides an extensive library of traffic blueprints including pedestrians, vehicles and street signs, etc. Many modern autonomous driving systems, e.g. Transfuser~\cite{Transfuser} and LAV~\cite{LAV}
%\textbf{LZ: add a few here,with citations? (Answered)}
, are developed with the CARLA simulator.

In this paper, the Scenario Runner is used to build common traffic scenarios in the simulator.
 Scenario Runner is a tool provided by CARLA to build traffic scenarios. The logic of a scenario is encoded into a behaviour tree, which is composed of non-leaf control nodes (Select, Sequence and Parallel) and leaf nodes (behaviours). A scenario is then executed following the state of its behaviour tree. 
\begin{figure}[t]
    \centering
        \centering
        \includegraphics[width=0.75\linewidth]{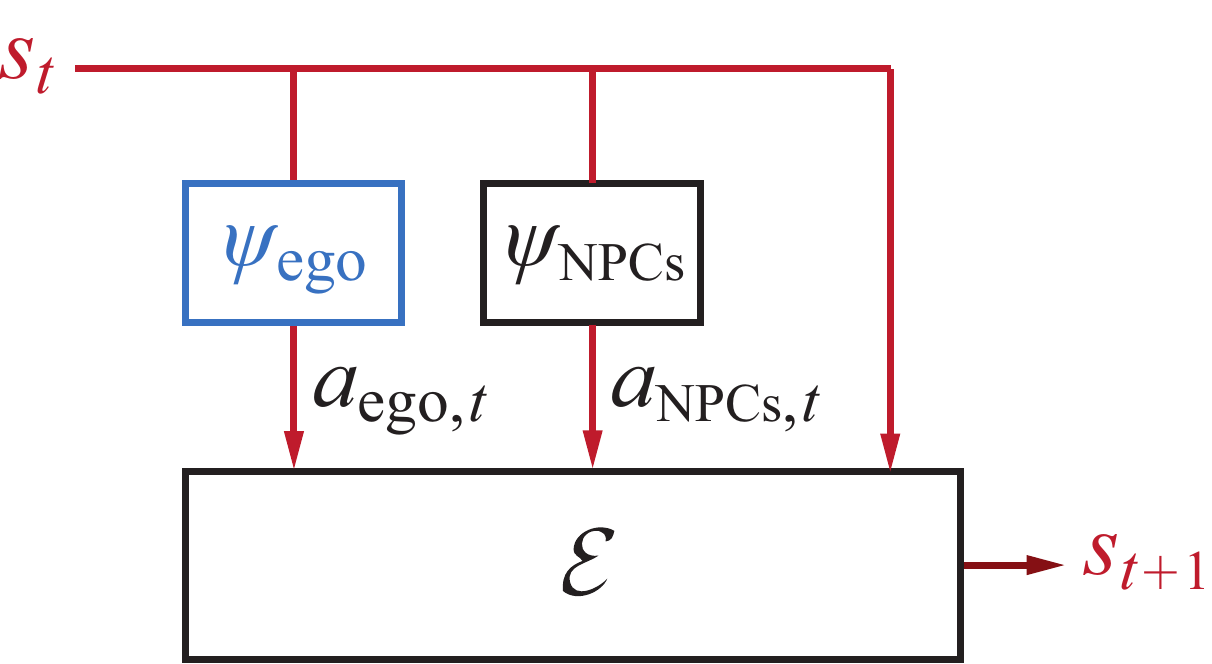}
    \caption{The simulator $\simulator$ iteratively generates the next state.}
    \label{fig:ADSmodel}
\end{figure}

\begin{figure}[t]
    \centering
        \centering
        \includegraphics[width=1\linewidth]{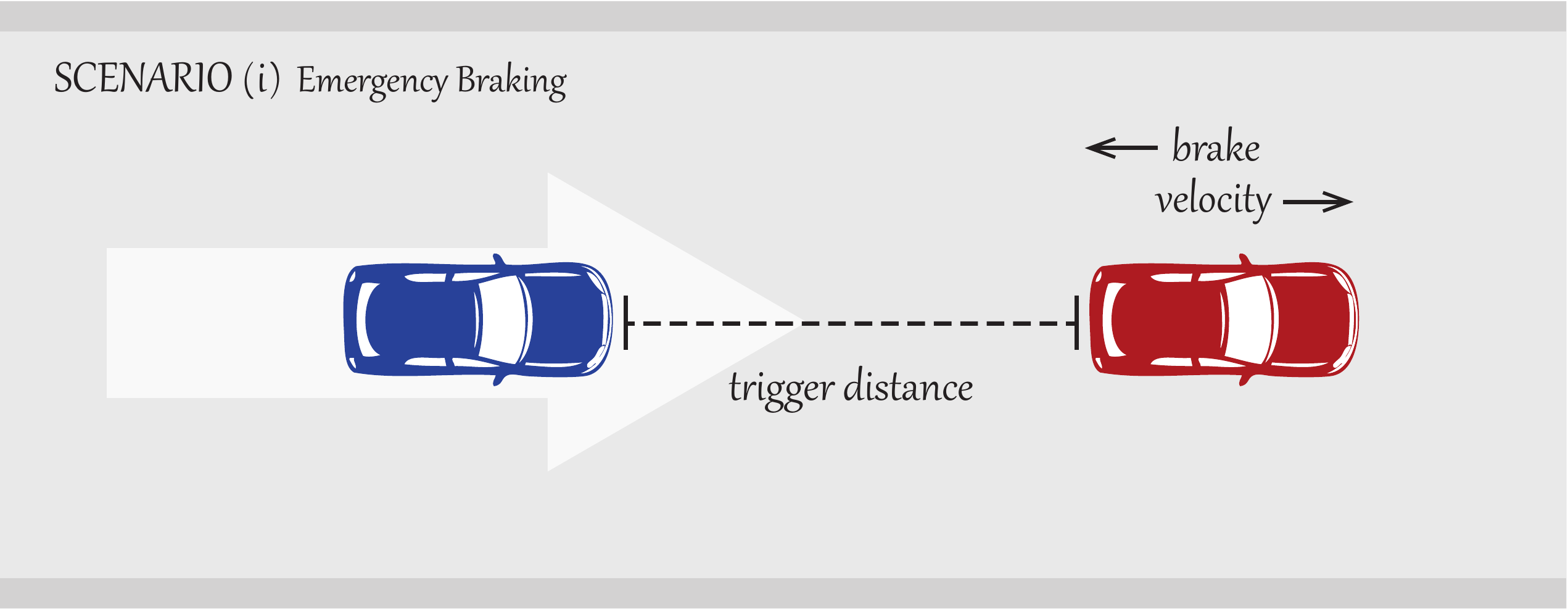}
    \caption{In this scenario, the ego vehicle drives along the road while the leading NPC vehicle brakes. Parameters are NPC agent speed $\theta_1$, the deceleration $\theta_2$, and the distance $\theta_3$ between two vehicles. 
    The behaviour model of the ego vehicle is its ADS, and the behaviour model $\psi_{\mathsf{NPC}_1}$ decelerates the NPC.  A function $\omega$ measures the distance between the ego vehicle and the leading NPC. The safety property is to guarantee the safe distance of $\tau = 0.1$(m) between two vehicles. 
    %The stop condition $\phi$ is that the safety property is violated or the simulation times out.
    }
    %The criteria model $\rho$ actually calculates the minimal distance between two vehicles in this scenario. Here, the threshold $\tau$ is set to represents the minimal safe distance between vehicles, e.g. $0.2$. We can instantiate an verification problem that checks whether the ADS can keep a safe distance ($\geq 0.2$) in all possible scenarios generated with different parameters.}
    \label{fig:example}
\end{figure}

\section{Scenario driven safety verification}
\label{sec:verification}
In this section, we first formalize the autonomous driving scenario (Section \ref{subsec:ads}). Then, we introduce a model learning based verification approach in Section \ref{sec:model_learning}. When the verification cannot conclude the safety property, in Section \ref{sec:parameter_space}, techniques are also proposed to partition the parameter space into safe/unsafe regions.

\subsection{Formalism of the autonomous driving scenario}
\label{subsec:ads}
An autonomous driving scenario comprises an autonomous driving agent $\ego$ and other agents $\NPC_1,\NPC_2,\ldots$ in the (simulator) environment.
Denote by $\Theta=[0,1]^m$ the set of the parameter vector $\theta=(\theta_1,....,\theta_m)$, where each $\theta_i$ is a normalized parameter variable of the scenario. 

We use $\mathcal{S}$ to denote the set of states. A state $s_t\in\mathcal{S}$ describes the  status of all agents in the virtual world at step $t$, including locations, speeds, accelerations, etc.
% \footnote{Obviously, the next state depends also on other environment parameters such as whether conditions, which are not the main focus of the paper, thus are skipped in the formal descriptions.}.
The behaviour of $\ego$ and $\NPC_i$ is modeled by the function $\psi_\ego(s_t)$ and $\psi_{\NPC_i}(s_t)$, 
which generate the action $a_{\ego,t}$ and $a_{\NPC_i,t}$ at step $t$, respectively. 

%\textbf{LZ: add set of parameters $\Theta$, and explain $\theta$}

% We use $s_t\in\mathcal{S}$ to denote the traffic state at step $t$.
% For a NPC, its behaviour is modeled via the function $\psi_i(s_t; \theta_{A_i})$ that generates the corresponding action $a_{i,t}$ using the current state $s_t$, where $\theta_{A_i}$ represents its parameters.
% The ego agent is built on various sensors, and it cannot directly access current state $s_t$. We define the \emph{observation} $o_t$ as the estimation of state $s_t$ given by the sensors in the ego vehicle. The behavioural model $\psi_{ego}(o_t;\theta_{ego})$ utilises the observation $o_t$ to produce the action of the ego vehicle $a_{ego, t}$ at step $t$.

The simulator $\simulator$ %$s_t;\theta_S)$ 
generates the next state $s_{t+1}$ according to the ego action $a_{\ego, t}$, the NPC actions $a_{\NPC_i,t}$, and the current state $s_t$ (see Figure~\ref{fig:ADSmodel}).
A simulation is a sequence of states $s_0, s_1, \ldots, s_{t_\bot}$ generated by simulator $\simulator$, %$S(\cdot;\theta_S)$ iteratively, 
where $s_0$ is the initial state and $s_{t_\bot}$ is the final state satisfying some terminating conditions.
% , subject to the scenario under consideration. 
Note that the initial state $s_0$ and the behaviour functions $\psi_\ego$ and $\psi_{\NPC_i}$ are all instantiated and fixed by the parameter $\theta$.
Therefore, the state sequence $s_0,s_1,\ldots,s_{t_{\bot}}$ is uniquely determined by the parameter $\theta$. Namely, each state $s_t$ at step $t$ can be considered as a function $s_t(\theta)$.
% can be inferred by a function $\bar{s}_t(\theta)$ with the parameter $\theta$.

%determined by the scenario~\footnote{Hereafter, we do not distinguish the initial state $s_0$ and its corresponding scenario.} 
%and $s_t$ is the first state satisfying the stop condition $\phi$.
% The stop condition $\phi:\mathcal{S} \rightarrow \{\top, \bot \}$ determines whether to stop the simulation at state $s$ i.e., a safety property is satisfied or not.\textbf{LZ: check where stop condition is used, probably canbe removed}

%To thoroughly test an ADS agent $A_{ego}$ in some scenario with NPCs $A_i$, testing approaches maneuver the parameter $\theta_i$, $\theta_{ego}$ and $\theta_S$ to create various test cases.
%For verification and testing purpose, an oracle $O(s):\mathcal{S} \rightarrow \{\top,\bot\}$ is implemented to decide whether a state meets stop condition i.e., a safety property is satisfied or not. 
%is violating the criterion. 
%We judge whether a simulation is a counter-example by evaluating all states in the simulation against the oracle, i.e., $\bigwedge_{1\leq i\leq t} O(s_i)$. Previous works introduce local search algorithms in the testing procedure, it exploits the results of prior test cases and helps generate new cases more likely to be a counter-example.

\subsubsection*{Safety properties}
\label{subsec:properties}
We are interested in the safety requirement of critical scenarios. In traffic scenarios, many safety properties can be described as a physical quantity (such as velocity, distance, angle, etc) always satisfying a certain threshold during the entire driving process.
In autonomous driving scenarios, we define a function $\omega$ to 
measure such physical quantity at a given state, and the safety properties can be defined as follows.
\begin{definition}[Safety Property]
For a given parameter $\theta\in\Theta$, a quantitative measure $\omega: \mathcal{S} \rightarrow \mathbb{R}$ and a threshold $\tau\in \mathbb{R}$, 
the state sequence $s_0, s_1, \ldots, s_{t_\bot}$ is safe if:
\begin{equation}\label{eq:criteria_model}
    \forall t\in\{0,1,\ldots,t_\bot\} \; \omega(s_t(\theta)) \ge \tau.
\end{equation}
\end{definition}
For simplicity, we introduce a \emph{fitness function}
$\rho(\theta)=\min_{0\leq i \leq t_\bot}{\omega(s_i(\theta))}$,
and the property can be equivalently rewritten as
$\rho(\theta) \geq \tau$.
For instance, we can use the distance between two vehicles as the quantitative measure $\omega$, and the collision free property requires that the distance is no smaller than a safe threshold $\tau > 0$.

The example in Figure \ref{fig:example} maps these notations above to a driving scenario. The question then is how to verify that an ADS meets the safe requirement in Equation (\ref{eq:criteria_model}), as a scenario can be initialised with all possible parameter values.

\subsection{Surrogate model based verification}
\label{sec:model_learning}

\begin{figure}[t]
    \centering
        \centering
        \includegraphics[width=0.8\linewidth]{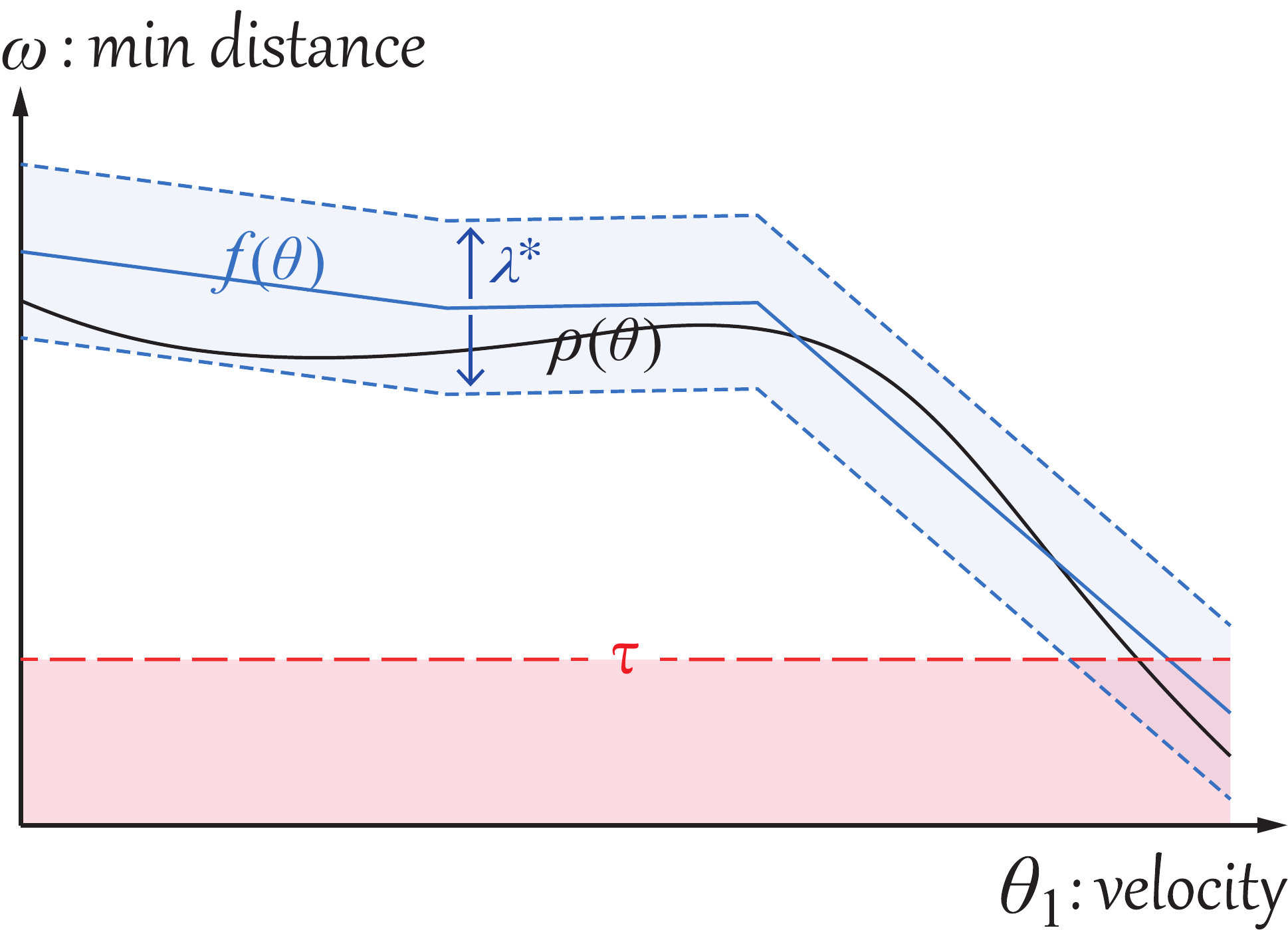}
    \caption{We plot the value of fitness function $\rho$ and the learned surrogate model $f$ with respect to the initial velocity $\theta_1$ of the NPC vehicle, while fixing other parameters. %We bound the surrogate model using the estimated 
    The fitness function $\rho$ is bounded by $f \pm \lambda^*$ with probabilistic guarantee. In this case, there exists initial velocity value that makes the distance lower bound from the surrogate model smaller than threshold $\tau$ (at bottom right corner), which violates Equation~\eqref{eq:verify_on_learned_model}. In other words, the ADS may break the collision free property.}
    \label{fig:examplemodel}
\end{figure}

\commentout{
\begin{figure}[t]
    \centering
        \centering
        \includegraphics[width=0.8\linewidth]{figures/framework2.eps}
    \caption{The xxxxxx}
    \label{fig:model_learning}
\end{figure}
}
%We have already formalise the verification problem into an optimization problem in previous section.
% \begin{equation*}
%     \max_{\theta \in \Theta}{\rho(\theta)}\\
%     s.t. s_j = S(a_{ego, j-1}, a_{i, j-1}, s_{i-1}; \theta_S) \forall 1 \leq j \leq t\\
% \end{equation*}
In general, it is intractable to check the safety property since the fitness function $\rho(\theta)$ depends on the behaviour models and the simulator, which cannot be expressed by explicit functions. Moreover, the simulator and the ADS often act in a black-box manner, which will further complicate the analysis. % in many situation, which makes it hard to analyse.
For verification purpose, we use a surrogate model $f$ (FNN with ReLU activate function in this paper) to approximate the original fitness function. An illustrative example is in Figure \ref{fig:examplemodel} for assisting the following discussion.

% By denoting the vector of the parameters as $\theta = (\theta_1, \ldots, \theta_m) \in \Theta$,
%Based on the surrogate model $f_n$ with $n$ steps of refinement, we try to verify the criteria on the it if we did not find any counter-examples in previous simulations. 
Solving the \emph{absolute distance} $\lambda$ between the surrogate model $f$ and the original fitness function $\rho$ can be encoded into an optimization problem:
\begin{equation}
    \label{eq:abs_dis_opt}
    \begin{split}
        & \min\limits_{\lambda \in \mathbb{R}}\lambda\\
        s.t.\;& \|f(\theta) - \rho(\theta)\| \leq \lambda,\;\forall \theta \in \Theta
    \end{split}
\end{equation}
%However, it is impossible to solve such optimization problem with infinite constraints. 
Based on the scenario optimization technique~\cite{scenario_theorem}, 
we reduce the infinite parameter set $\Theta$ to a finite subset $\Theta_{\mathrm{sample}}$ containing $K$ samples.
%We then use the scenario optimization technique~\cite{} to reduce the optimization problem by only considering a subset of the parameters $\Theta_0$ containing $K$ samples according to a probability measure $\mathbb{P}$.
\begin{equation}
    \label{eq:scenario_opt}
    \begin{split}
        & \min\limits_{\lambda \in \mathbb{R}}\lambda\\
        s.t.\;& \|f(\theta) - \rho(\theta)\| \leq \lambda,\;\forall \theta \in \Theta_{\mathrm{sample}}
    \end{split}
\end{equation}
As a result, the absolute distance given by Equation~\eqref{eq:scenario_opt} can be guaranteed with an error rate $\epsilon$ and a significance level $\eta$ according to the following lemma.
\begin{lemma}[see \cite{scenario_theorem}]\label{scenariotheorem}
    If \eqref{eq:scenario_opt} is feasible and has a unique optimal solution $\lambda^*$, and
    \begin{equation}
        \label{eq:scenariotheorem}
        \epsilon\geq \frac{2}{K}(\ln\frac{1}{\eta}+1),
    \end{equation}
    where $\epsilon$ and $\eta$ are the pre-defined error rate and the significance level, respectively, then with confidence at least $1-\eta$, the optimal $\lambda^*$ satisfies all the constraints in $\Omega$ but only at most a fraction of probability measure $\epsilon$, i.e., $\mathbb P (\|f(\theta) - \rho(\theta)\| > \lambda^*)\leq \epsilon$.
\end{lemma}

\paragraph{Neural network verification} After we have the absolute distance evaluation $\lambda^*$, one can use existing neural network verification tools like DeepPoly~\cite{deeppoly} and MILP~\cite{sherlock} to determine whether
\begin{equation}
    \label{eq:verify_on_learned_model}
    f(\theta) - \lambda^* \geq \tau, \forall \theta \in \Theta
\end{equation}
where $f(\theta) - \lambda^*$ serves as a probabilistic lower bound of the original model's fitness function $\rho$, and we remind that $\tau$ is the threshold for safety requirements.
%Here, we actually evaluate the probabilistic upper bound $f_n(\theta) + \lambda^*$ of the criteria model $\rho$.. 
Finally, when Equation~\eqref{eq:verify_on_learned_model} holds, the verification returns SAFE, and we can conclude that the autonomous driving system satisfies the safety property with error rate $\epsilon$ and significance level $\eta$:
\begin{equation*}
    \begin{split}
        \mathbb{P}(\rho(\theta)\geq \tau) &\geq \mathbb{P}(\rho(\theta)\geq f(\theta) - \lambda^*)\\
        &\geq \mathbb{P}(\|f(\theta) - \rho(\theta)\| \leq \lambda^*) \geq 1-\epsilon.
    \end{split}
\end{equation*}
In summary, by given enough samples $\Theta_{\mathrm{sample}}$, the verification results on a learned surrogate model $f$ can be transferred to the original fitness function $\rho$ with guarantee.
 
%Previous works~\cite{} proposed a learning-based method to approximate the complex model with a surrogate model. In this paper, we also adopt the idea to learn a surrogate DNN model from the criteria model $\rho$. We first use random fuzzer to sample $N_0$ initial parameters $\theta^{(0)}$ from $\theta$ and run the simulator to achieve the simulation results and then analyse the criteria quantification. A initial DNN model $f_0(\theta) : \Theta \rightarrow \mathbb{R}$ are trained using these $N_0$ samples, which approximates the original criteria model $\rho(\theta)$.

\paragraph{Surrogate model learning} As mentioned above, we adopt model learning to approximate, with the guarantee in Lemma \ref{scenariotheorem}, the original fitness function. 
% The idea is to develop a surrogate model to predict generating a sequence of states that are as close as possible to the original traffic scenario's, guided by the fitness function $\rho$.
% In the model learning, we train the surrogate by a randomly sampled parameter set $\hat{\Theta}$. 
Here, we present our learning procedure depicted in Algorithm~\ref{algo:model_learning}. 
The major part of the algorithm is a while loop, in which the surrogate model is trained iteratively with the parameter set $\hat\Theta$ (and the corresponding simulation outcome).
After each iteration of training, the absolute distance evaluation $\lambda^*$ is calculated (in Line 5),  
and the learning procedure terminates when $\lambda^*$ is small enough. 
Otherwise, the model is not sufficiently trained, 
and new parameters will be added into the parameter set (see Line 9--11) for the next training step. 
In addition to the incrementally sampled parameters $\Theta_\mathrm{inc}$,
we further derive the extreme parameters (denoted by $\Theta_\mathrm{ex}$) that potentially deviate far from the fitness function at current step, where the surrogate model has minimum and maximum outputs.
These parameter samples can be generated by using the attack algorithms (such as PGD~\cite{PGD}) with maximization and minimization directions.
Finally, the surrogate model is obtained, as well as the distance $\lambda^*$ from the fitness function guaranteed with the error rate $\epsilon$ and the significance level $\eta$.

\begin{algorithm}[t]
  \caption{Surrogate Model Learning}
  \label{algo:model_learning}
  \begin{flushleft}
    \textbf{INPUT:}\,\, A driving scenario, error rate $\epsilon$, significance level $\eta$ \\
    \textbf{OUTPUT:}\,\, Surrogate model $f$, Absolute distance $\lambda^*$
  \end{flushleft}
  \begin{algorithmic}[1]
    \STATE initialise a surrogate model $f$
    \STATE sample a parameter set $\hat\Theta\subset \Theta$
    % \STATE $\hat{\Theta}\leftarrow \mbox{sample } N_0 \mbox{ parameter vectors from } \Theta$
    %\STATE $train\_data\leftarrow\{\}$
    %\FOR{$\theta\in\hat{\theta}$}
    %        \STATE $train\_data\leftarrow train\_data\cup\{\simulator(\theta)\}$
    %\ENDFOR
    \WHILE {\emph{the time threshold is not met }}
        \STATE train $f$ with all $\theta\in\hat{\Theta}$ and  $\rho(\theta)$.
        % \STATE $f\leftarrow \mbox{ train the surrogate model with } \hat{\Theta}$
        \STATE $\lambda^* \leftarrow \mbox{the solution of the optimization problem~\eqref{eq:scenario_opt}}$
        \IF {$\lambda^*$ is small enough}
            \STATE \textbf{break}
        \ELSE
            \STATE incrementally sample the parameters $\Theta_\mathrm{inc}\subset\Theta$
            % \STATE $\hat{\Theta}_{\mathrm{clean}} \leftarrow \{\mbox{new random parameters}\}$
            \STATE generate the extreme parameters $\Theta_\mathrm{ex}\subset\Theta$
            % \STATE $\hat{\Theta}_{\mathrm{adv}} \leftarrow \{\mbox{new extreme parameters}\}$
            \STATE $\hat{\Theta}\leftarrow \hat{\Theta}\cup {\Theta}_{\mathrm{inc}} \cup {\Theta}_{\mathrm{ex}}$
        \ENDIF
    \ENDWHILE
    \RETURN $f$, $\lambda^*$
  \end{algorithmic}
\end{algorithm}

\subsection{Parameter space exploration}
\label{sec:parameter_space}
However, since autonomous driving systems are complex combinations of many components and algorithms, it is hard for them to stay safe in the whole parameter space.  
When the verification does not return SAFE, it is meaningful to further analyse the relationship between the unsafe behaviour and the parameters, which will provide an important reference for improving the system.
Thus, based on the parameters we care about, which we called \emph{associated parameters}, we divide the parameter space into cells, and in a quantitative way, an indicator $\varrho\in[0,+\infty]$ can be computed to express how unsafe the model is within each cell.

Based on two associated parameters, saying $\theta_1$ and $\theta_2$, we can split the two-dimensional parameter space into a $l$-by-$l$ grid where each square has the same length $\delta=1/l$. Namely, the whole parameter space is divided into $l^2$ cells as $\Theta=\bigcup_{i,j=0,\ldots,l-1} \Theta_{i,j}$ where
\[
\Theta_{i,j}=[i\delta,(i+1)\delta]\times[j\delta,(j+1)\delta]\times[0,1]^{m-2}.
\]
Then, for the cell $\Theta_{i,j}$, we can define the quantitative unsafe indicator $\varrho_{i,j}$ as its minimal value satisfying
\[
\forall \theta \in \Theta_{i,j} \; f(\theta) \ge \tau-\varrho_{i,j}.
\]
Note that the quantified formula can be encoded by linear constraints, so we can compute $\varrho_{i,j}$ by MILP.
Intuitively, each $\tau-\varrho_{i,j}$ indicates the maximal threshold such that the surrogate model is safe with all $\theta\in\Theta_{i,j}$.
So, the region
\[
\Theta_{\mathrm{safe}}=\bigcup_{\varrho_{i,j}=0}\Theta_{i,j}
\]
is exactly an under-approximation of the parameter region that the surrogate is safe, and a larger $\varrho_{i,j}$ implies that the ADS is more prone to unsafe behaviour in such scenario within the corresponding parameter region.
In this work, we  focus on the analysis for two associated parameters since the results can be easily visualised by heat map, but it is straightforward to generalise this analysis to more associated parameters.

\begin{figure*}[!htb]
  \centering
  \subfloat{
    \includegraphics[width=1\linewidth]{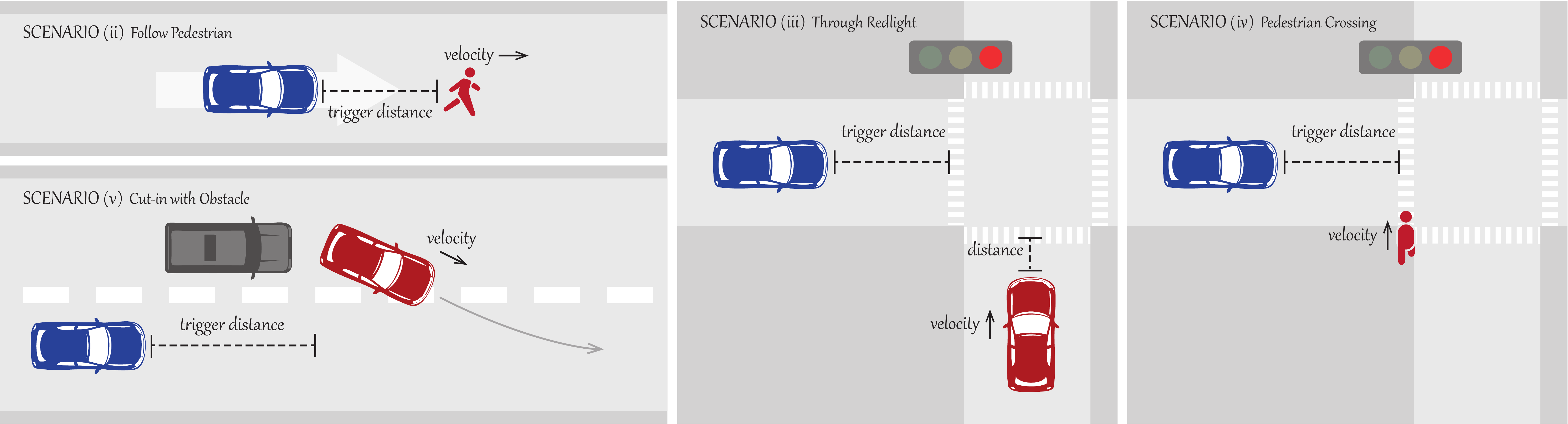}
  }
  \caption{(ii) \emph{Follow Pedestrian:} In this scenario, the ego car keeps a safe distance with the pedestrian in front. %There parameters in this scenario are the NPC velocity, the initial distance and the threshold distance.
  (iii) \emph{Through Redlight:} The ego car encounters a NPC car running the red light when crossing the junctions. %The NPC car will be spawned when the distance between the ego car and the junction is smaller than a threshold.
  %Three parameters are the NPC velocity, the distance between the NPC car and the junction, and the threshold distance.
  (iv) \emph{Pedestrian Crossing:} A pedestrian crosses the road while the ego car enters the junction. %The pedestrian will be spawned when the distance between the ego car and the junction is smaller than a threshold. The two parameters in this scenario are the pedestrian velocity and the threshold distance.
  (v) \emph{Cut-in with Obstacle:} In this scenario, a NPC car in front of a van tries to cut into the road where the ego car drives along. %The NPC performs the cut-in when the distance between it and the ego car is smaller than a threshold. The three parameters in this scenario are the NPC velocity, the initial distance between the ego car and the NPC car, and the threshold distance.
  }
  \label{fig:scenarios}
\end{figure*}

\section{Experiments}
\label{sec:experiment}

In this section, we apply the proposed verification in Section \ref{sec:verification} together with the state-of-the-art autonomous driving system LAV \cite{LAV} \footnote{LAV is ranked the 3rd place in the CARLA Learderboard but the first two systems are not publicly available before submission.}. The safety requirement is to assure a safe road distance between the ego agent and the NPCs in various scenarios.

\paragraph*{Setup} Five traffic scenarios in Figure~\ref{fig:example} and Figure~\ref{fig:scenarios} are considered. There are two variants for each scenario at different locations, labelled with ``Case \#1'' and ``Case \#2''. 
Besides similar parameters as detailed in Figure~\ref{fig:example}, there are also 8 weather parameters in each scenario, including cloudness, fog density, precipitation, precipitation deposits, sun altitude angle, sun azimuth angle, wetness and wind intensity.
Furthermore, an additional parameter is introduced to control the initial distance between the ego vehicle and the NPC vehicle.
The minimal safe distance between agents is $\tau = 0.1$ for all the scenarios. 
%We evaluate each scenario on two different roads, which generate two variants with similar traffic semantics.

The surrogate model is a 3-layer FNN with 100 neurons in each hidden layer.
The error rate $\epsilon$ and the significance level $\eta$ of the surrogate model are $0.01$ and $0.001$ respectively. 
% In other words, we want to verify LAV is $99\%$ safe with $99.9\%$ confidence, at least.
The number of the initial learning samples is $900$, and we train the surrogate model in $10$ iterations of refinement by increasing $40$ random samples and $10$ extreme samples after each iteration.
All the experiments are conducted on a server with AMD EYPC 7543 CPU, 128G RAM and 4 Nvidia RTX 3090.

\subsection{Discovery of ADS glitch} %Discover Abnormal Behaviours of LAV}
% We first demonstrate how the proposed verification algorithm can be used to debug the LAV autonomous driving system.
Before the verification, our framework can help discover functional glitches for the LAV autonomous driving system.
% -->>>Our algorithm outputs the absolute distance evaluation $\lambda^*$ between the surrogate model $f$ and the fitness function $\rho$ by solving the optimization problem~\eqref{eq:scenario_opt} mentioned in Sect.~\ref{sec:model_learning}. 
We find that sometimes the absolute distance evaluation $\lambda^*$ between the surrogate model $f$ and the fitness function $\rho$ is very large, for instance in the \emph{Follow Pedestrian} scenario and \emph{Through Redlight } scenario. 
Such unusual $\lambda^*$ is an indication of an abnormal ADS behaviour.
% there are more than $3\,000$ runs used to train the surrogate model and evaluate the absolute distance. We set a threshold of absolute distance being $12$ and $10$ for the follow pedestrian scenario and the through redlight scenario, respectively. Then, we go through the results, and identify  $1$  and $22$ runs in the two  scenarios, respectively. 
By outlier detection and checking the corresponding simulation records, we identify $1$ and $26$ abnormal cases in $3\,000$ runs for the two scenarios, respectively.
These cases expose a functional glitch of the LAV system: \emph{the ego car can randomly halt in the middle of the road for no reason}, resulting in an extremely large distance. 

To eliminate the influence of the glitch, we have to discard the abnormal cases. After the surrogate models are retrained, the absolute distance $\lambda^*$ is reduced sharply from $13.4436$ to $0.7087$ in the Follow Pedestrian scenario, and from $12.0283$ to $2.6678$ in the Through Redlight scenario.

\subsection{Safety verification}

We first verify whether the ADS can always keep  a safe distance in the studied scenarios. The results are reported in Table~\ref{tab:verification_results}.
%The safe distance threshold $\tau$ varies in different scenarios as the size and orientation of the vehicles and pedestrians are different.
The safe distance threshold $\tau$ is set to $0.1$ for all scenarios.
The verification concludes that one emergency braking scenario and two Follow Pedestrian scenarios (out of ten) are safe.

While the results in Table~\ref{tab:verification_results} indicate that certain scenarios (e.g., when NPCs are in front of the ego vehicle) are more safe than the others (e.g.,  when NPCs intrude from the sides), it is not surprising that there exist road hazards that result in safety violation in the verification, given the rich parameter conditions. This fact confirms the importance of the parameter space exploration phase.

\begin{table}[!htb]
\renewcommand{\arraystretch}{1.2}
\caption{Safety verification results of different scenarios: $\lambda^*$ is the resulting absolute difference. In the ``Counter-example.'' column, the checkmark (\checkmark) means the ADS is safe regarding to the scenario, otherwise an adversarial output of the surrogate model is given.}
\centering
\begin{tabular}{ccccc}
%\toprule
\hline %\rule{0pt}{10pt}
                       Scenario    & Case & $\lambda^*$ & Counter-example  \\ \hline %\midrule
\multirow{2}{*}{Emergency Braking}    & \#1 & 6.8699      & 5.4876    \\ %\cline{2-2} \cline{4-5} 
                                      & \#2 &                    5.4194      & \checkmark \\ \hline
\multirow{2}{*}{Follow Pedstrain}     & \#1 & 0.7087      & \checkmark \\ %\cline{2-2} \cline{4-5} 
                                      & \#2 &                   4.2193      & \checkmark \\ \hline
\multirow{2}{*}{Through Redlight}     & \#1 & 4.6885      & -5.8308   \\ %\cline{2-2} \cline{4-5} 
                                      & \#2 &                   2.9944      & -5.0789   \\ \hline
\multirow{2}{*}{Pedestrian Crossing}  & \#1 & 2.6678      & -0.5953    \\ %\cline{2-2} \cline{4-5} 
                                      & \#2 &                    2.8866      & -1.7328   \\ \hline
\multirow{2}{*}{Cut-in with Obstacle} & \#1 & 0.4004      & -0.5991    \\ %\cline{2-2} \cline{4-5} 
                                      & \#2 &                    0.8790      & 0.0772    \\   \hline %\bottomrule
\end{tabular}
\label{tab:verification_results}
\end{table}

\begin{figure}[t]
    \centering
        \centering
        \includegraphics[width=0.99\linewidth]{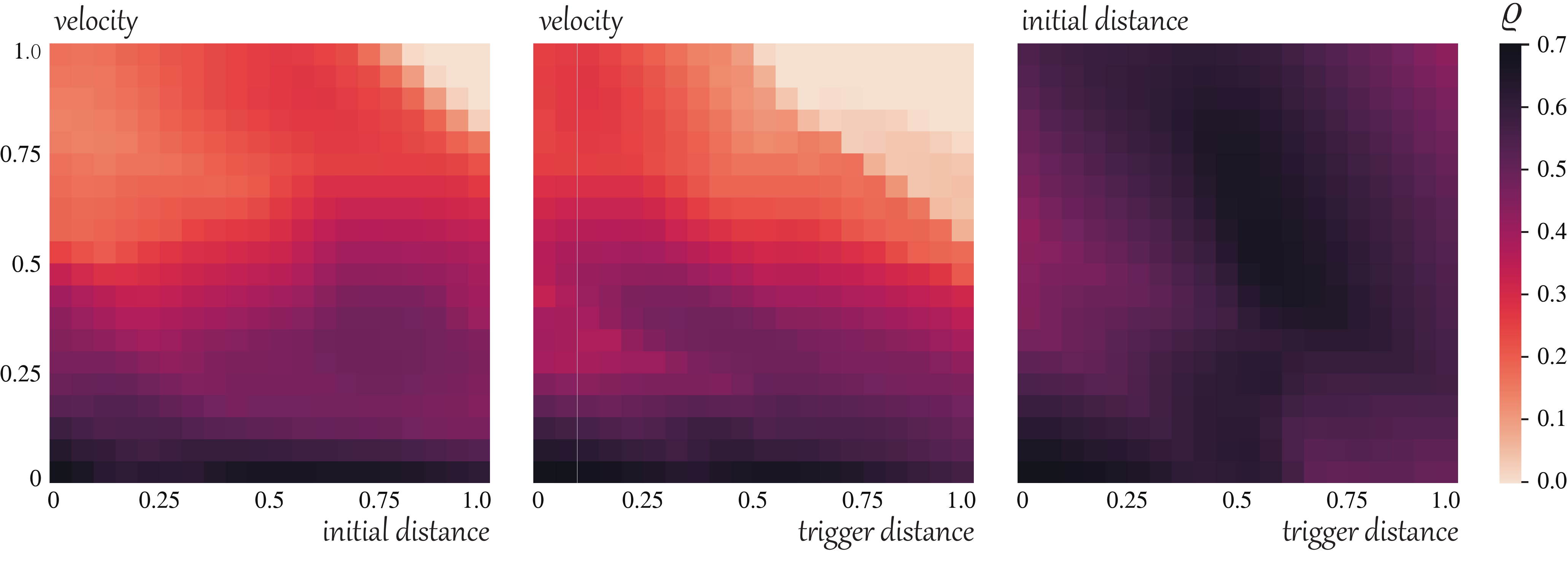}
    \caption{By heatmap, the results of parameter space exploration is illustrated for the cut-in scenario \#1. The grid marked with darker color implies the ADS is more likely to violate the safety property with the parameters in it.
    }
    \label{fig:Unsafe_Range_Cutin_With_Obstacle}
\end{figure}

\begin{figure}[t]
    \centering
        \centering
        \includegraphics[width=0.9\linewidth]{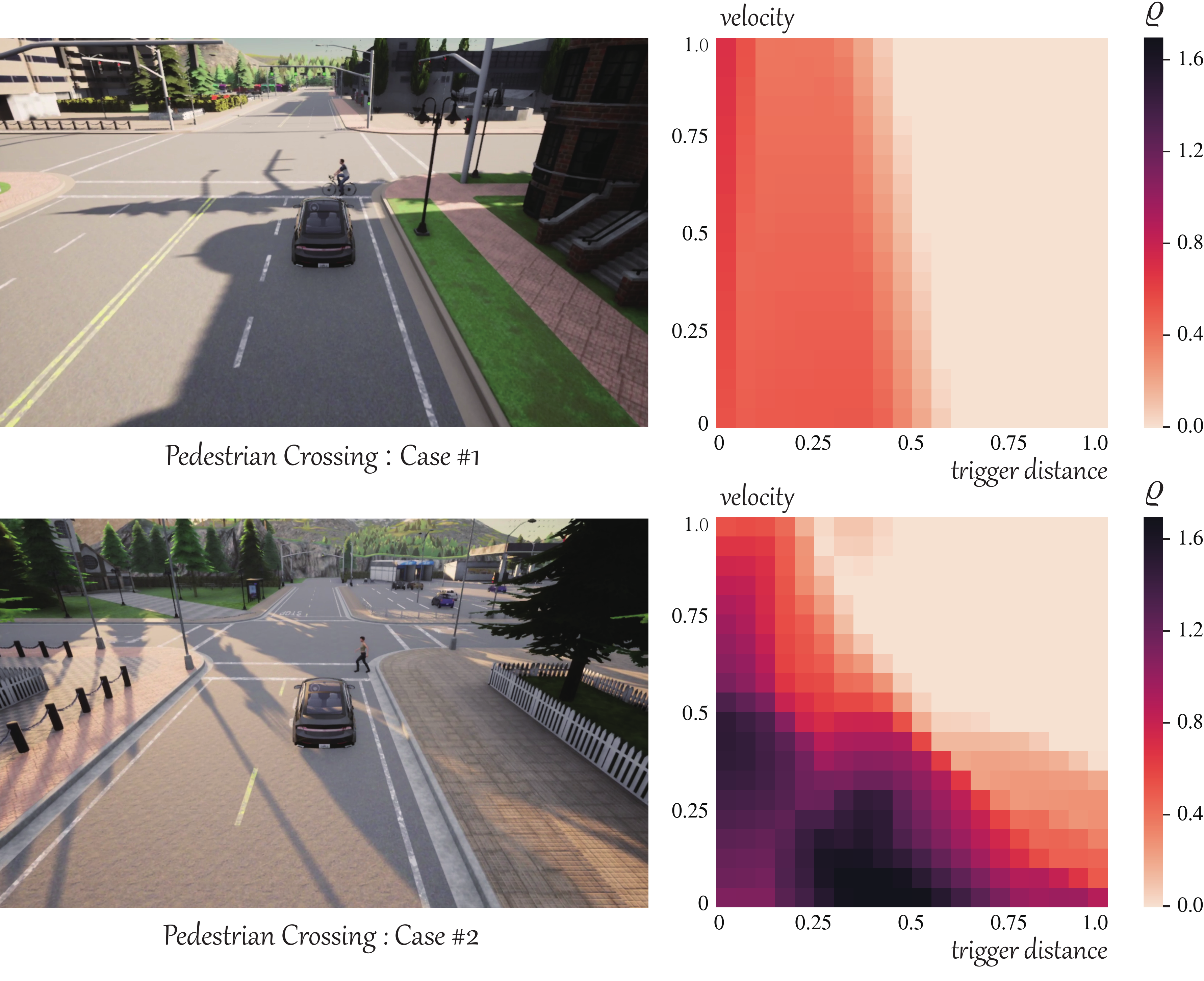}
    \caption{Parameter space exploration is applied on the pedestrian crossing scenarios. The heatmaps show obviously different characteristics for Case~\#1 (with a cyclist) and Case~\#2 (with a walking pedestrian).}
    \label{fig:Unsafe_Range_Pedestrian_Crossing}
\end{figure}

\subsection{ADS behaviour analysis}
% \begin{figure}[t]
%     \centering
%         \centering
%         \includegraphics[width=1\linewidth]{cvpr2023-author_kit-v1_1-1/latex/figures/Figure_Emergency_Braking.pdf}
%     \caption{Unsafe parameter range of the emergency braking scenario in location \#1.}
%     \label{fig:Unsafe_Range_Emergency_Braking}
% \end{figure}

In our approach, parameter space exploration  is conducted (see Sect.~\ref{sec:parameter_space}) to analyse the behaviour of LAV, following driving hazards identified via the verification.  
For two associated parameters, we divide the range $[0,1]$ evenly into $20$ intervals, and the quantitative indicator $\varrho_{i,j}$ for all parameter cells can be calculated. 
Here, we analyse a) the cut-in scenario \#1 with three different parameters and b) the Pedestrian Crossing scenario in two different cases with the same parameters.

\emph{a) }
For the cut-in scenario \#1, we focus on three associated parameters---initial distance, trigger distance, and NPC velocity. 
The analysis result is illustrated in Figure~\ref{fig:Unsafe_Range_Cutin_With_Obstacle}.
From the left most figure and the middle figure, it is clear that the ADS is more likely to keep the safe distance of the ego vehicle when the NPC velocity is high and the initial distance and trigger distance are large enough. This is reasonable as a faster NPC car and a longer distance leave more room for the ADS to overtake and make a braking response to avoid collision.

%Comparing the first two figure, we found that they share some similar characteristics, which shows the two distance parameters are somehow correlated. It is quite intuitive since the two parameters both characterize the distance between the ego vehicle and the NPC vehicle.

%The first impression on the results is that the unsafe indicator is much larger 
Interestingly, the unsafe indicator values are much larger in the right most figure, and it means that by the two distance parameters alone, the initial distance and the trigger distance, it is unable to reach a safe condition. %well depict the safety property and not very relative to the safety property. 
In other words, the proper NPC velocity parameter value is required for any safe traffic scenario.

%is more important and needs to be carefully evaluated for further improvements.

% 1. velocity is a more relevant parameter
% 2. the effect on initial distance and trigger distance are similar, larger distance can produce a safer scenario
% 3. larger the speed of the pedestrian and trigger distance, safer the scenario.. Insights..

\emph{b) }
Here, we report the quantitative unsafe indicator for two Pedestrian Crossing cases regards the NPC velocity and trigger distance in Figure~\ref{fig:Unsafe_Range_Pedestrian_Crossing}.
In the Pedestrian Crossing scenario \#1, we spawn a cyclist instead of a walking pedestrian, and the trigger distance is larger. It needs to be mentioned that the cyclist are faster (2-3 m/s) than the walking pedestrian (1-2 m/s). 

From Figure~\ref{fig:Unsafe_Range_Pedestrian_Crossing}, we can conclude that larger NPC velocity and larger trigger distance make both scenarios safer for the ADS. Bedsides, the smaller unsafe indicator values also imply that the scenario \#1 are even safer than the scenario \#2. These meet the intuition that larger velocity and longer trigger distance can make the ADS more likely to sense the moving pedestrian. It is worth noting that the velocity parameter in first scenario (which is faster) seems less relative to the safety property than in the second scenario, and this captures the different behaviours of a cyclist and walking pedestrian. % which suggests us use a walking pedestrian instead of a cyclist in further evaluation.

\section{Conclusion and future work}
In this paper, we propose a novel approach to verify the safety property of a given ADS with a probabilistic guarantee. % in traffic scenarios. 
The safety property is quantified by a fitness function with scenario parameters.  A surrogate model is learned for approximating this fitness function under the traffic scenario, and the safety property verified in the surrogate model can be transferred to the original ADS with specified confidence and error rate. 
When the verification fails, the parameter space is partitioned into safe and unsafe regions with quantitative indicators. This parameter space exploration further helps the testing and improvement of autonomous driving under different scenarios. The experiments validate the utility of our approach with promising results and vivid examples.
%ADS fails the verification, we use the unsafe quantitative indicator to analyse the parameters, which can extract insights from the surrogate model and help explain the behaviour of the ADS. In experiment, we evaluate the ADS LAV to verify whether it can keep a safe distance in some common traffic scenarios. The results show that our approach is a practical method to evaluate modern ADS.

As for future work, we believe that the verification driven surrogate model learning will be an important direction for AI safety and security in general.
%plan to embed verification into learning procedure to improve the surrogate by utilising the intermediate verification results. 
We are also interested in applying the parameter analysis in this work to the ``design space exploration'' of autonomous driving systems. %proposing safety guidelines for ADS based on our approach.

% \section*{Acknowledgment}

% The preferred spelling of the word ``acknowledgment'' in America is without 
% an ``e'' after the ``g''. Avoid the stilted expression ``one of us (R. B. 
% G.) thanks $\ldots$''. Instead, try ``R. B. G. thanks$\ldots$''. Put sponsor 
% acknowledgments in the unnumbered footnote on the first page.

% \section*{References}

\bibliographystyle{IEEEtran}
\bibliography{IEEEabrv,egbib}

\end{document}